\documentclass[10pt,twocolumn,letterpaper]{article}

\usepackage[pagenumbers]{cvpr} 

\usepackage{tabularx}
\usepackage{colortbl} %
\usepackage{xspace}

\usepackage{graphicx,booktabs,amsmath,amssymb,mathtools,multirow,tikz}
\usetikzlibrary{arrows.meta,positioning,fit}
\usepackage{pgfplots}
\pgfplotsset{compat=1.18}

\usepackage{xspace}
\definecolor{cvprblue}{rgb}{0.21,0.49,0.74}
\usepackage[pagebackref,breaklinks,colorlinks,allcolors=cvprblue]{hyperref}

\def\confName{CVPR}
\def\confYear{2027}

\title{Fast-JEPA: Advancing the Success–Latency Frontier for Latent World Models}

\author{%
	Yao Zhang\textsuperscript{1,2,8}\quad
	Bin Yu\textsuperscript{3,2,8}\quad
	Shijie Lian\textsuperscript{4,2,8}\quad
	Haishan Liu\textsuperscript{5,2,8}\quad
	Changti Wu\textsuperscript{6,2,8}\\
	Yu Wu\textsuperscript{1,$\dagger$}\quad
	Kai Chen\textsuperscript{2,7,8,$\dagger$}\\[4pt]
	{
		\textsuperscript{1}Wuhan University\quad
		\textsuperscript{2}Zhongguancun Academy
	}\\
	{
		\textsuperscript{3}Harbin Institute of Technology\quad
		\textsuperscript{4}Huazhong University of Science and Technology
	}\\
	{
		\textsuperscript{5}Nanjing University\quad
		\textsuperscript{6}East China Normal University
	}\\
	{
		\textsuperscript{7}Zhongguancun Institute of Artificial Intelligence\quad
		\textsuperscript{8}DeepCybo
	}
}

\begin{document}
\raggedbottom
\brokenpenalty=10000
\maketitle

\begingroup
\renewcommand{\thefootnote}{\ensuremath{\dagger}}
\footnotetext[0]{Corresponding author.}
\endgroup

\begingroup
\renewcommand{\thefootnote}{\ensuremath{}}
\footnotetext[0]{Work done during internship at DeepCybo.}
\endgroup

\begin{abstract}
Latent world models predict the outcomes of action sequences, but comparing candidate plans through recursive rollout remains expensive at test time. We introduce \textbf{Fast-JEPA}, a latent world model planning framework that advances the success–latency Pareto frontier of predictive search. Fast-JEPA reorganizes predictive computation to support effective action-sequence search at low planning cost. It constructs a goal-directed plan to guide search and directly predicts the outcomes of complete candidates for efficient comparison, avoiding recursive state rollout during refinement. Across PushT, Reacher, Cube, and TwoRoom, Fast-JEPA achieves 96.50\% average success rate with $2.41\times$ faster planning than INTACT. Search-budget sweeps show that its competitive success and low latency extend across candidate counts and refinement depths. We further introduce \textbf{DynaGoal}, a benchmark for evaluating world-model planning under dynamic out-of-distribution conditions, execution deadlines, and controlled response delays. Fast-JEPA achieves average success rates of 73.83\% on DynaGoal-Shift and 87.67\% on DynaGoal-Deadline, demonstrating the practical value of low-latency predictive planning under changing control conditions and execution-time constraints. Our project page is available at \href{https://deepcybo-physai.github.io/Fast-JEPA/}{https://deepcybo-physai.github.io/Fast-JEPA/}.
\end{abstract}

\section{Introduction}
\label{sec:intro}
World models enable visual controllers to compare action
sequences through predicted outcomes in representation
space~\cite{hafner2019planet,maes2026lewm}.
However, evaluating many candidate sequences through
predictive search can incur substantial online computation.
This raises a central question: how can world-model planning
advance the success--latency Pareto frontier while retaining
the benefits of predictive search?

A complete planner must decide both which actions are worth considering and how much computation to spend comparing them. A strong initial plan reduces the burden on search, but it does not eliminate the potential value of comparing alternatives. Under a limited online budget, refinement must explore useful candidates while controlling the cost of their evaluation. With recursive scoring, each candidate requires a chain of intermediate state predictions, and this computation is repeated across search rounds.

\begin{figure}[!tp]
	\centering
	\includegraphics[width=\linewidth]{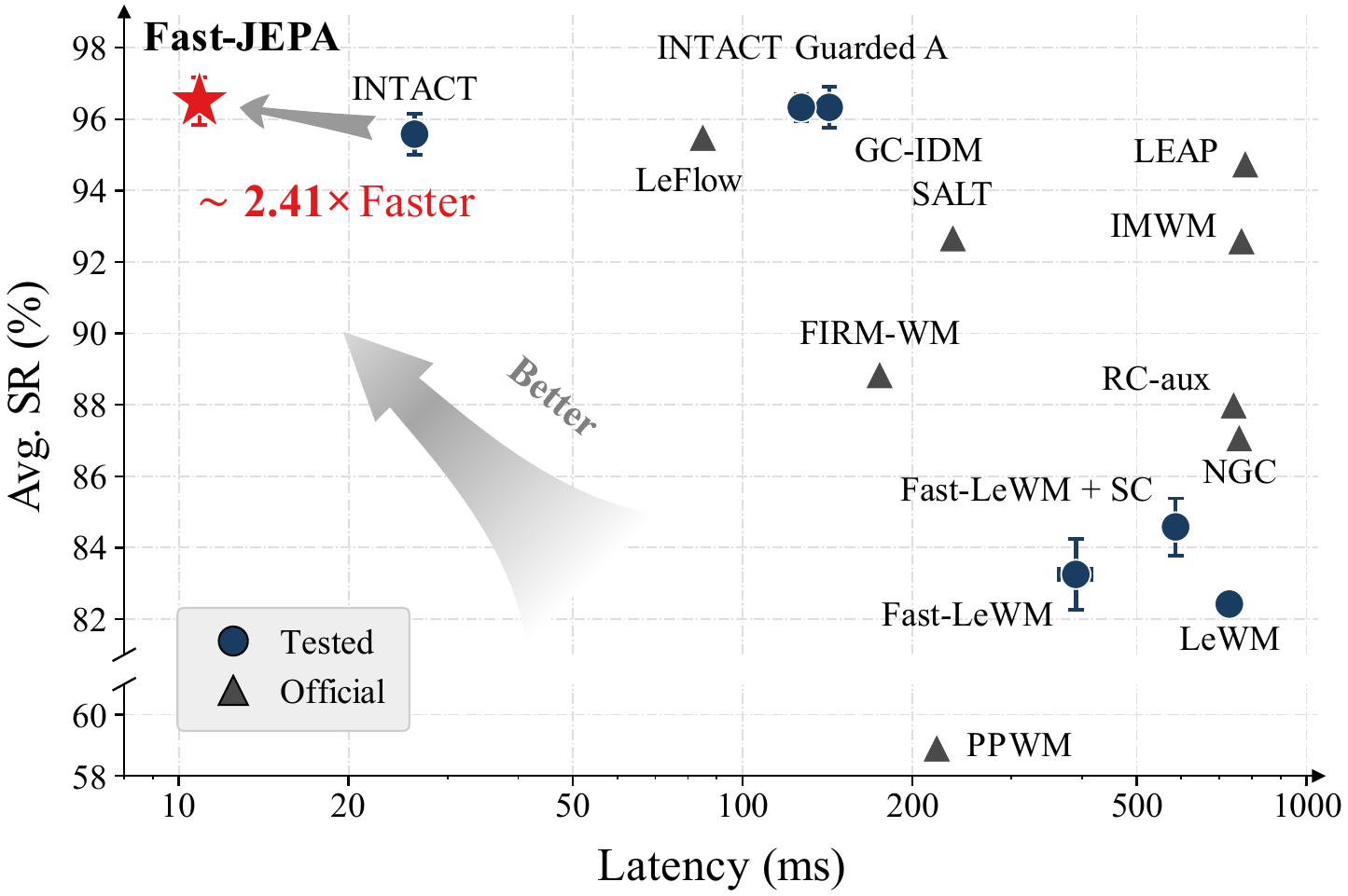}
	\caption{\textbf{Advancing the success--latency Pareto frontier.} Fast-JEPA achieves 96.50\% average success rate at 10.88\,ms, improving the observed frontier among the locally evaluated configurations and running $2.41\times$ faster than INTACT.}
	\label{fast}
\end{figure}

Action construction and candidate comparison expose different information to the planner. During sequential construction, later actions remain unspecified, so intermediate outcome predictions provide context for subsequent action choices. Once a candidate is complete, all its actions are known. Under a terminal visual-goal objective, comparison requires only its predicted endpoint. This distinction motivates retaining intermediate queries for action construction and using direct terminal queries for candidate evaluation.

\begin{figure*}[!t]
\centering
\includegraphics[width=\linewidth]{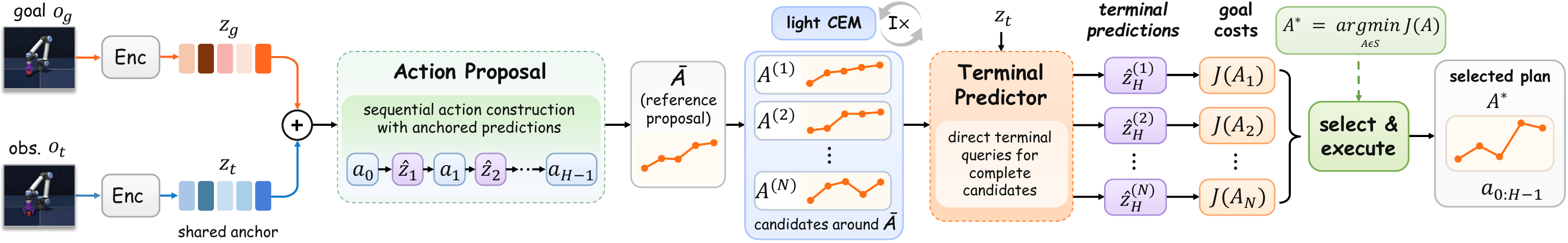}
\caption{\textbf{Fast-JEPA planning overview.}
Current and goal representations condition sequential proposal
construction, where intermediate outcome queries guide subsequent
action chunks. Local CEM samples alternatives around $\bar A$,
and batched terminal queries provide goal costs for selecting
$A^\star$. Both intermediate and terminal predictions use the
same forward world model anchored to the current observed
representation $z_t$. The $+$ symbol denotes conditioning.
The proposal and best-so-far plan remain eligible during search.
After selection, $L$ primitive actions are executed before
replanning from a new observation.}
\label{fig:pipeline}
\end{figure*}

We introduce \textbf{Fast-JEPA}, a planning framework that reorganizes predictive computation to support action-sequence search at low planning cost (Figure~\ref{fig:pipeline}). Fast-JEPA constructs a goal-directed plan through world-model-guided action generation, providing a starting point for predictive search. Complete candidate sequences are then compared through direct terminal predictions under a shared visual-goal objective, avoiding recursive state rollout during refinement. By concentrating search around an informed plan while reducing repeated candidate-evaluation costs, Fast-JEPA retains the benefits of predictive search within a substantially lower computational budget.

To evaluate world-model planning under dynamic out-of-distribution (OOD) conditions and temporal constraints, we introduce \textbf{DynaGoal}, a benchmark comprising three complementary protocols across PushT, Reacher, Cube, and TwoRoom. DynaGoal-Shift evaluates adaptation to changing goals and environment geometry, while DynaGoal-Deadline measures goal completion under joint planning and execution time budgets. DynaGoal-Delay further examines robustness to controlled response delays as environments evolve between observation and action.

Across PushT, Reacher, Cube, and TwoRoom, Fast-JEPA
achieves 96.50\% average success rate with a mean
complete planning latency of 10.88\,ms. This search-based controller plans $2.41\times$ faster than INTACT while attaining higher average success (Figure~\ref{fast}). Search-budget sweeps show that its low-latency operating region extends across candidate counts and refinement depths, and the scoring comparison links lower planning cost to cheaper candidate evaluation. Fast-JEPA also achieves 73.83\% average success on DynaGoal-Shift and 87.67\% on DynaGoal-Deadline. Together, these results demonstrate that Fast-JEPA maintains competitive success at substantially lower planning latency.

Our contributions can be summarized as follows:
\begin{itemize}
	\setlength{\itemsep}{2pt}
	\setlength{\parskip}{0pt}
	\setlength{\parsep}{0pt}
	\item We present \textbf{Fast-JEPA}, a world-model planning
	framework that advances the success--latency Pareto
	frontier by reorganizing predictive computation for
	efficient action-sequence search.
	\item We introduce \textbf{DynaGoal}, a benchmark
	evaluating world-model planning under
	out-of-distribution conditions, execution deadlines,
	and response delays.
	\item Extensive experiments demonstrate that Fast-JEPA advances the success--latency Pareto frontier of world-model planning across diverse control tasks.
\end{itemize}

\section{Related Work}

\noindent\textbf{Latent world models.}
Visual foresight and latent dynamics enable action selection through predicted
outcomes~\cite{ebert2018visual,hafner2019planet,chua2018pets}.
Dreamer learns behavior through imagined trajectories~\cite{hafner2020dreamer,hafner2025dreamerv3},
while TD-MPC couples latent planning with learned policies and value
estimates~\cite{hansen2022tdmpc,hansen2024tdmpc2}.
Predictive representation learning extends this direction from images to
video~\cite{assran2023ijepa,bardes2024vjepa,assran2025vjepa2}.
LeWM learns visual latent dynamics~\cite{maes2026lewm}, while Fast-LeWM improves prediction efficiency through structural design~\cite{gao2026fast}. FIRM-WM studies state-factorized recurrent dynamics for visual planning,
while RC-aux introduces multi-horizon prediction and reachability
supervision~\cite{wu2026firmwmstatefactorizedfactualinterventionalrecurrent,li2026predictiveplannablercauxlatent}.
Together, these works develop predictive representations and dynamics
for control through different temporal prediction structures.

\noindent\textbf{Goal-directed action learning and planning.}
Goal-conditioned learning supplies policies for reaching desired
states~\cite{andrychowicz2017her,nair2018rig,ghosh2021gcsl}.
Action-chunk and diffusion policies generate candidate behavior through
sequence modeling~\cite{zhao2023act,chi2023diffusionpolicy,janner2022diffuser}.
GC-IDM amortizes control into inverse dynamics~\cite{nguyen2026latent},
while INTACT studies intent-to-action learning and reference-guided
search for latent control~\cite{sun2026intact}.
Learned priors guide planning in POPLIN~\cite{wang2019poplin},
and sampling-based improvements such as iCEM improve search
efficiency~\cite{pinneri2021icem}.
LeFlow generates latent paths and verifies their decoded actions,
while LEAP refines goal-conditioned proposals through differentiable
rollouts~\cite{huang2026leflowgenerativelatentflow,pham2026latentenergyactionplanning}.
IMWM combines retrieval initialization with hybrid
scoring~\cite{gao2026imwmintuitionmodelscomplement}, whereas NGC
uses neighborhood graphs to define latent-space goal
distances~\cite{you2026latentworldmodelsplan}.
Together, these approaches illustrate different ways to guide
candidate generation, evaluation, and selection in predictive control.

\section{Fast-JEPA}
\label{sec:method}

Fast-JEPA organizes world-model queries according to the information required during search (Figure~\ref{fig:pipeline}). During action construction, outcomes predicted from the observed anchor under the selected actions inform subsequent action choices. Once an action sequence is complete, its predicted terminal outcome is sufficient for comparison under the visual-goal objective. The framework retains sequential dependence in action construction while avoiding recursive state prediction during candidate evaluation.

\subsection{Problem Formulation}

At decision time $t$, an image encoder represents the observation $o_t$ and
goal image $o_g$ in a common latent space,
\begin{equation}
  z_t=E_\theta(o_t),\quad z_g=E_\theta(o_g).
\end{equation}
An action plan $A=(a_0,\ldots,a_{H-1})\in
\mathbb R^{H\times B\times d_u}$ contains $H$ chunks, each with $B$
primitive actions of dimension $d_u$. We write
$A_{:k}=(a_0,\ldots,a_{k-1})$ for its first $k$ chunks. Unsupervised
visual representations make it possible to compare predicted and desired
outcomes in latent space~\cite{caron2021dino,oquab2024dinov2}.

The planning objective is to select an action sequence whose predicted terminal representation is close to the visual goal. The prediction horizon $HB$ and the feedback interval $L$ are distinct: the controller executes $L$ primitive actions from a selected plan before receiving a new observation and replanning. We focus on the online computation required to select complete action sequences under this terminal visual-goal
objective.

\subsection{Overall Framework}

To evaluate both partial and complete action sequences within a single planning process, Fast-JEPA uses a shared forward world model. The predictor is accessed through an anchored outcome query,
\begin{equation}
  \mathcal Q_\psi(z_t,A,k)
  \triangleq W_\psi(z_t,A_{:k}),\quad 1\leq k\leq H.
  \label{eq:anchored_query}
\end{equation}
The query takes the observed representation $z_t$, rather than a previously
predicted state, as its anchor. The prediction horizon $k$ identifies the requested
outcome. It is not an additional learned model input. In particular,
$\mathcal Q_\psi(z_t,A,k)$ supplies intermediate predictions and
$\mathcal Q_\psi(z_t,A,H)$ supplies the terminal prediction of a complete
plan. The notation $\mathcal Q_\psi$ denotes the existing forward world
model, not a separate prediction network.

A planning cycle first constructs a goal-directed proposal using
intermediate outcome queries, then evaluates complete alternatives
through direct terminal predictions. The resulting search is
centered on the proposal, which remains eligible for final
selection. Throughout a solve, the observed anchor $z_t$ and
model parameters remain fixed.

\subsection{Prediction-Guided Plan Construction}

The construction process must choose actions that are not yet specified. We
represent its partial plan by the pair
$\xi_k=(\bar A_{:k},\hat z_k)$, where $\bar A_{:k}$ records the selected
chunks and $\hat z_k$ is their outcome predicted from the observed anchor $z_t$. Starting with
$\bar A_{:0}=\varnothing$ and $\hat z_0=z_t$, the partial plan is extended sequentially by
\begin{equation}
\begin{aligned}
  \bar a_k &= G_\phi(\hat z_k,z_g,a_{k-1}^{\mathrm{hist}}),\\
  \bar A_{:k+1} &= \bar A_{:k}\oplus\bar a_k,\\
  \hat z_{k+1} &= \mathcal Q_\psi(z_t,\bar A_{:k+1},k+1),
                    \quad k<H-1.
\end{aligned}
\label{eq:proposal_recursion}
\end{equation}
The first two operations apply for $0\leq k<H$. The third is needed only
for $0\leq k<H-1$. Here $\oplus$ denotes concatenation,
$G_\phi$ is the conditional mean of the learned action distribution, and
$a_{k-1}^{\mathrm{hist}}=\bar a_{k-1}$ when $k>0$. At $k=0$ it is the
latest available chunk from $h_t$. After all $H$ chunks are constructed, the resulting
proposal is $\bar A=\bar A_{:H}$.
This proposal provides a goal-directed starting point
for subsequent predictive refinement.

Equation~\eqref{eq:proposal_recursion} specifies a sequential action dependency, not a recursive state-transition model. An anchored outcome prediction conditioned on the selected
actions guides the next chunk, but every world-model call predicts from
$z_t$ and the accumulated actions. In particular, the predicted outcome $\hat z_k$ is never
substituted for $z_t$ as the anchor of any subsequent
world-model query.

The final action chunk requires no additional intermediate
query. Construction therefore entails $H$ action-generation
steps and only $H-1$ anchored outcome queries.
The resulting sequence initializes the search distribution
and remains eligible for selection under the same goal
objective used to evaluate subsequent candidates.

\subsection{Terminal-Guided Predictive Search}

Once a candidate is complete, its full action sequence is known. For the
terminal visual-goal objective, selection requires the predicted endpoint,
not a chain of intermediate predictions. We accordingly define
\begin{equation}
\begin{aligned}
  J(A,z_t,z_g)
    &= \big\|\mathcal Q_\psi(z_t,A,H)-z_g\big\|_2^2,\\
  A^\star &= \underset{A\in\mathcal S}{\arg\min}\;J(A,z_t,z_g),
\end{aligned}
\label{eq:cost}
\end{equation}
where $\mathcal S$ contains all evaluated candidates, including the original
proposal and the final mean candidate. Complete candidates are scored by
batched direct terminal queries from the same observed anchor, without
recursively feeding intermediate predicted states back into the world model.

To allocate additional search around the constructed plan, we use
proposal-centered cross-entropy method (CEM) updates
~\cite{rubinstein1999cem}. The initial search distribution in the flattened
action space is
\begin{equation}
  q^{(0)}=\mathcal N(\mu^{(0)},\operatorname{diag}(v^{(0)})),
  \quad \mu^{(0)}=\bar A,\quad v^{(0)}=\sigma_0^2\mathbf 1.
  \label{eq:search_init}
\end{equation}
At refinement round $j$, two of the $N$ candidate slots are reserved for
$\bar A$ and the best plan found so far. The other $N-2$ are sampled from
$q^{(j)}$. The best-so-far plan is initialized to $\bar A$. After terminal
scoring, the $K$ lowest-cost slots define elite indices $\mathcal E_j$. The search distribution is updated using their coordinatewise moments,
\begin{equation}
\begin{aligned}
  \mu^{(j+1)} &= \frac{1}{K}\sum_{i\in\mathcal E_j} A_i^{(j)},\\
  \widetilde v^{(j+1)}
    &=\frac{1}{K}\sum_{i\in\mathcal E_j}
      \big(A_i^{(j)}-\mu^{(j+1)}\big)^{\odot 2}.
\end{aligned}
\label{eq:elite_moments}
\end{equation}
The updated variance $v^{(j+1)}$ is obtained with the standard-deviation
bounds given in Appendix~\ref{sec:implementation}. After $I$ rounds, the
final distribution mean $\mu^{(I)}$ is also scored. The controller executes
the lowest-cost evaluated plan. Retaining the original proposal guarantees
that refinement cannot worsen its predicted goal cost, although this
does not guarantee improved realized success in actual execution.

\subsection{Learning and Computational Properties}

\noindent\textbf{Learning anchored outcomes.}
For a training segment $\tau\sim\mathcal D$, let
$z_k^\tau=E_\theta(o_{t+kB})$, let $A_{:k}^\tau$ be the
first $k$ recorded action chunks, and let $z_0^\tau$ denote the
initial observed representation.

The forward world model is supervised at multiple horizons from a shared
observed anchor, with each prediction conditioned on the corresponding
action sequence:
\begin{equation}
  \mathcal L_{\mathrm{wm}}
  =\mathbb E_{\tau\sim\mathcal D}\!\left[
    \frac{1}{Hd_z}\sum_{k=1}^{H}
    \big\|
      \mathcal Q_\psi(z_0^\tau,A^\tau,k)
      -z_k^\tau
    \big\|_2^2
  \right],
  \label{eq:wm_loss}
\end{equation}
where $d_z$ is the latent dimension.

Each target is predicted from the same observed anchor,
allowing intermediate and terminal queries to share
the learned model.

\noindent\textbf{Learning goal-conditioned actions.}
The conditional action model parameterizes a diagonal
Gaussian $p_\phi(a\mid c)$, whose mean is used for plan
construction.

The action model is trained with two conditioning targets: the immediately following latent state and the segment endpoint.
For each demonstrated action chunk, we define
\begin{equation}
  \Delta_k^r =
  \begin{cases}
    z_{k+1}^\tau-z_k^\tau,
      & r=\mathrm{next},\\
    \operatorname{sg}(z_{\mathrm{end}}^\tau)-z_k^\tau,
      & r=\mathrm{end},
  \end{cases}
  \label{eq:action_conditions}
\end{equation}
where $\operatorname{sg}$ denotes stop-gradient. The corresponding action condition is
$c_k^r=(z_k^\tau,\Delta_k^r,a_{k-1}^\tau)$.

Both target spans supervise the same demonstrated
action chunk through a shared conditional distribution:
\begin{equation}
  \mathcal L_{\mathrm{act}}
  =-\frac{1}{Bd_u}\mathbb E_{\tau,k}
  \left[
    \sum_{r\in\{\mathrm{next},\mathrm{end}\}}
    \lambda_r
    \log p_\phi(a_k^\tau\mid c_k^r)
  \right].
  \label{eq:action_loss}
\end{equation}
The initial preceding action chunk is taken from the
recorded causal history. The full joint objective,
conditioning features, action normalization, and
implementation details are provided in
Appendix~\ref{sec:implementation}.

\noindent\textbf{Computational analysis.}
Let $N$ denote the number of candidates per CEM
round, $I$ the number of refinement rounds, and
$H$ the action-chunk horizon. Let $G$ and $W$
denote the computation required by one action-generator
evaluation and one world-model prediction, respectively.

A rollout-based CEM planner requires $H$
world-model predictions per candidate. In contrast,
Fast-JEPA constructs an initial proposal using $H$
action-generator evaluations and $H-1$ anchored
outcome predictions, followed by direct terminal
evaluation of complete candidates. Including one
final mean-candidate evaluation for both CEM
variants, the principal model computation is
\begin{equation}
\begin{gathered}
\begin{aligned}
C_{\mathrm{CEM\text{-}rollout}}
&=(NI+1)HW,\\[3pt]
C_{\mathrm{Fast\text{-}JEPA}}
&=
\underbrace{HG+(H-1)W}_{\substack{\text{Proposal}\,\text{construction}}}
+
\underbrace{(NI+1)W}_{\substack{\text{Terminal}\,\text{evaluation}}}\\[3pt]
&=HG+(H+NI)W,
\end{aligned}
\\[7pt]
\mathcal O(NIHW)
\;\longrightarrow\;
\mathcal O((H+NI)W).
\end{gathered}
\label{eq:prediction_calls}
\end{equation}
The asymptotic expression assumes $G=O(W)$ and
comparable costs for individual world-model predictions.
Actual planning latency also depends on batching and
CEM overhead, and is evaluated empirically.

\begin{figure}[!t]
\centering
\includegraphics[width=\linewidth]{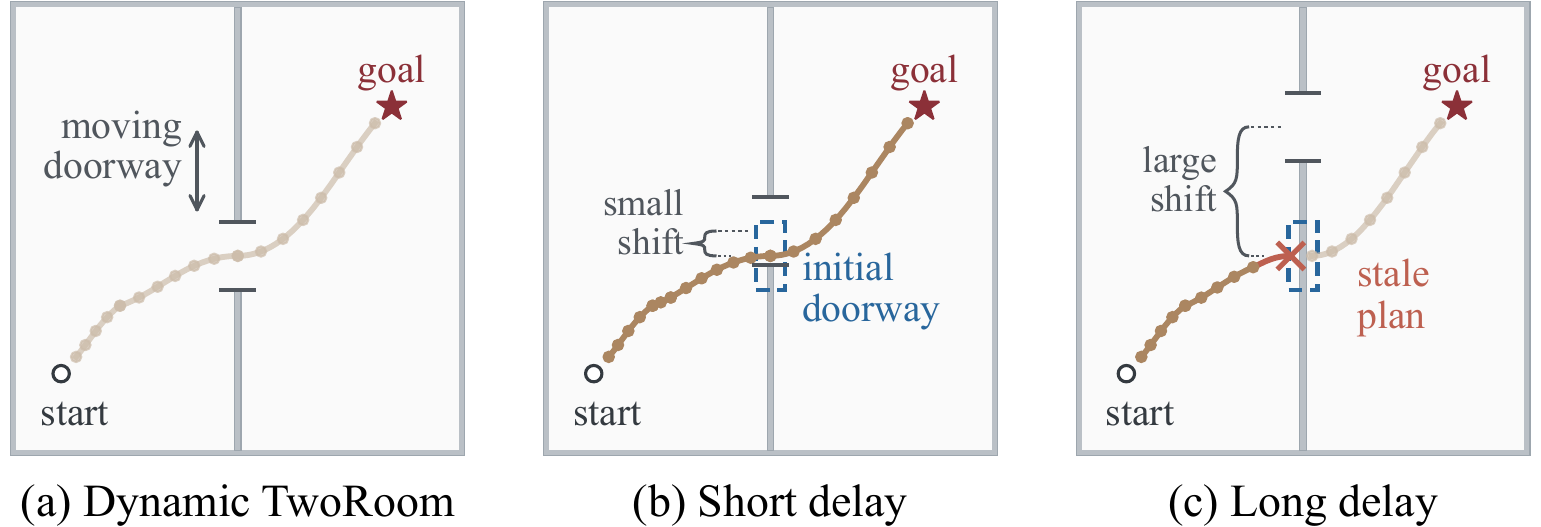}
\caption{\textbf{Dynamic TwoRoom and response delay.} (a) A continuously moving doorway. (b,c) Alternative short- and long-delay outcomes from the same initial observation. Blue outlines mark the initial doorway. Trajectories are schematic and doorway displacement is enlarged for clarity.}
\label{fig:tworoom_delay_schematic}
\end{figure}

\begin{table*}[!t]
\centering\footnotesize
\setlength{\tabcolsep}{2.5pt}
\renewcommand{\arraystretch}{1.16}
\caption{\textbf{Four-task success and planning latency.} SR is in percent and latency in ms. Bold marks the best local aggregate results.}
\label{tab:main_results}
\begin{tabularx}{\textwidth}{@{}l*{6}{>{\centering\arraybackslash}X}@{}}
\toprule
Method & PushT & Reacher & Cube & TwoRoom & Avg. SR (\%) & Latency (ms)\\
\midrule
\multicolumn{7}{l}{\textit{Published results}}\\
\rowcolor{gray!10}
IMWM~\cite{gao2026imwmintuitionmodelscomplement} & $92.70$ & $83.80$ & $94.70$ & $99.20$ & $92.60$ & $767$\\
\rowcolor{gray!1}
NGC~\cite{you2026latentworldmodelsplan} & $91.20$ & $99.40$ & $57.90$ & $99.80$ & $87.08$ & $760$\\
\rowcolor{gray!10}
LEAP~\cite{pham2026latentenergyactionplanning} & $90.00$ & $89.00$ & $100.00$ & $100.00$ & $94.75$ & $779$\\
\rowcolor{gray!1}
FIRM-WM~\cite{wu2026firmwmstatefactorizedfactualinterventionalrecurrent} & $75.70$ & $92.70$ & $88.00$ & $99.00$ & $88.85$ & $175$\\
\rowcolor{gray!10}
LeFlow~\cite{huang2026leflowgenerativelatentflow} & $95.20$ & $86.80$ & $100.00$ & $100.00$ & $95.50$ & $85$\\
\rowcolor{gray!1}
RC-aux~\cite{li2026predictiveplannablercauxlatent} & $90.80$ & $87.20$ & $76.00$ & $98.00$ & $88.00$ & $743$\\
\rowcolor{gray!10}
SALT~\cite{zhang2026onestepaccuracystateaffinelatent} & $88.00$ & $90.70$ & $94.00$ & $98.00$ & $92.68$ & $236$\\
\rowcolor{gray!1}
PPWM~\cite{feng2026parallelpredictiveworldmodels} & $37.67$ & $97.00$ & $45.67$ & $55.33$ & $58.92$ & $221$\\
\midrule
\multicolumn{7}{l}{\textit{Locally evaluated reference configurations}}\\
\rowcolor{gray!10}
LeWM~\cite{maes2026lewm} & $90.67\!\pm\!0.58$ & $86.33\!\pm\!2.52$ & $65.00\!\pm\!1.00$ & $87.67\!\pm\!2.52$ & $82.42\!\pm\!0.14$ & $729.94\!\pm\!1.61$\\
\rowcolor{gray!1}
GC-IDM~\cite{nguyen2026latent} & $86.00\!\pm\!2.00$ & $100.00\!\pm\!0.00$ & $99.33\!\pm\!1.15$ & $100.00\!\pm\!0.00$ & $96.33\!\pm\!0.58$ & $142.41\!\pm\!1.50$$^{*}$\\
\rowcolor{gray!10}
Fast-LeWM~\cite{gao2026fast} & $90.33\!\pm\!1.53$ & $83.33\!\pm\!3.21$ & $69.67\!\pm\!3.06$ & $89.67\!\pm\!2.31$ & $83.25\!\pm\!1.00$ & $389.99\!\pm\!25.93$\\
\rowcolor{gray!1}
Fast-LeWM + SC~\cite{gao2026fast} & $92.00\!\pm\!2.00$ & $84.67\!\pm\!2.08$ & $75.00\!\pm\!3.00$ & $86.67\!\pm\!2.31$ & $84.58\!\pm\!0.80$ & $586.15\!\pm\!29.06$\\
\rowcolor{gray!10}
INTACT~\cite{sun2026intact} & $87.33\!\pm\!2.31$ & $96.00\!\pm\!0.00$ & $100.00\!\pm\!0.00$ & $99.00\!\pm\!0.00$ & $95.58\!\pm\!0.58$ & $26.19\!\pm\!0.22$\\
\rowcolor{gray!1}
INTACT Guarded A~\cite{sun2026intact} & $91.33\!\pm\!1.53$ & $97.00\!\pm\!0.00$ & $99.67\!\pm\!0.58$ & $97.33\!\pm\!0.58$ & $96.33\!\pm\!0.38$ & $127.09\!\pm\!2.81$\\
\rowcolor{blue!10}
\textbf{Fast-JEPA (Ours)} & $90.00\!\pm\!3.00$ & $96.33\!\pm\!1.15$ & $100.00\!\pm\!0.00$ & $99.67\!\pm\!0.58$ & $\mathbf{96.50\!\pm\!0.66}$ & $\mathbf{10.88\!\pm\!0.04}$\\
\bottomrule
\end{tabularx}
\end{table*}

\section{DynaGoal Benchmark}
\label{sec:dynagoal_design}

Online control must accommodate evolving environments
while allocating time to planning and execution.
We introduce \textbf{DynaGoal}, a benchmark comprising
three complementary control task families across PushT,
Reacher, Cube, and TwoRoom: Shift evaluates adaptation
to changing control conditions, Deadline tests
time-constrained goal completion, and Delay examines
robustness to delayed planning responses.
All tasks reuse models trained on the original
environments without additional training or fine-tuning.

\noindent\textbf{DynaGoal-Shift.}
These tasks introduce changing targets or environment
geometry during control. PushT uses a drifting target
that reverses direction. Reacher and Cube switch targets,
and TwoRoom contains a continuously moving doorway
(Figure~\ref{fig:tworoom_delay_schematic}(a)).
Success requires sustained goal reaching for PushT
and TwoRoom, and completion before and after the target
change for Reacher and Cube.
Measured planning delay determines when a new plan
becomes available, directly connecting planning cost
to control performance in continuously evolving
environments.

\noindent\textbf{DynaGoal-Deadline.}
These tasks require goal completion within task-specific
deadlines that account for both cumulative planning time
and action execution duration.
Success therefore depends on the planner's ability
to complete the task within the strict time budget
by balancing planning and execution time.
Appendix~\ref{sec:seven_extensions} specifies the
Shift and Deadline task conditions, success criteria,
and timing rules.

\noindent\textbf{DynaGoal-Delay.}
These evaluations extend the four Shift tasks with
controlled total response delays of 0, 300, 500,
and 800\,ms.
They assess robustness as the interval between
observation and plan execution increases
(Figure~\ref{fig:tworoom_delay_schematic}(b,c)).
During this interval, the environment advances using
buffered actions from the previous plan, holding the
last action when the buffer is empty.
The new plan replaces the remaining buffered actions
once available.
The imposed delay sets the response interval
independently of measured planning time.
Each controller is evaluated on 100 episodes per task
at each delay level using the Shift success criteria.

\begin{table*}[!t]
\centering\footnotesize
\setlength{\tabcolsep}{2.5pt}
\renewcommand{\arraystretch}{1.15}
\caption{\textbf{DynaGoal-Shift results.} SR is in percent and average planning latency in ms. Protocols appear in Appendix~\ref{sec:seven_extensions}.}
\label{tab:extension_results}
\begin{tabularx}{\textwidth}{@{}l*{6}{>{\centering\arraybackslash}X}@{}}
\toprule
Method & PushT-shift & Reacher-shift & Cube-shift & TwoRoom-shift & Avg. SR (\%) & Latency (ms) \\
\midrule
\rowcolor{gray!10}
LeWM~\cite{maes2026lewm} & $0.00\!\pm\!0.00$ & $2.00\!\pm\!1.00$ & $1.67\!\pm\!0.58$ & $3.33\!\pm\!2.08$ & $1.75\!\pm\!0.66$ & $767.51\!\pm\!1.72$\\
\rowcolor{gray!1}
Fast-LeWM~\cite{gao2026fast} & $1.33\!\pm\!2.31$ & $5.33\!\pm\!2.08$ & $4.00\!\pm\!1.00$ & $11.67\!\pm\!2.52$ & $5.58\!\pm\!0.95$ & $404.43\!\pm\!5.57$\\
\rowcolor{gray!10}
Fast-LeWM + SC~\cite{gao2026fast} & $0.67\!\pm\!0.58$ & $3.33\!\pm\!2.08$ & $2.33\!\pm\!1.53$ & $14.33\!\pm\!5.51$ & $5.17\!\pm\!1.84$ & $618.62\!\pm\!12.95$\\
\rowcolor{gray!1}
GC-IDM$^{*}$~\cite{nguyen2026latent} & $1.33\!\pm\!2.31$ & $7.00\!\pm\!2.00$ & $8.00\!\pm\!3.00$ & $38.67\!\pm\!3.21$ & $13.75\!\pm\!1.25$ & $170.42\!\pm\!3.44$\\
\rowcolor{gray!10}
INTACT~\cite{sun2026intact} & $50.67\!\pm\!2.08$ & $73.00\!\pm\!2.65$ & $56.00\!\pm\!1.73$ & $45.33\!\pm\!5.03$ & $56.25\!\pm\!1.64$ & $24.62\!\pm\!0.35$\\
\rowcolor{gray!1}
INTACT Guarded A~\cite{sun2026intact} & $36.33\!\pm\!6.35$ & $48.67\!\pm\!3.06$ & $15.00\!\pm\!1.00$ & $41.00\!\pm\!4.36$ & $35.25\!\pm\!1.56$ & $115.43\!\pm\!1.61$\\
\rowcolor{blue!10}
\textbf{Fast-JEPA (Ours)} & $57.33\!\pm\!8.62$ & $80.67\!\pm\!4.16$ & $72.00\!\pm\!3.00$ & $85.33\!\pm\!5.13$ & $\mathbf{73.83\!\pm\!4.16}$ & $\mathbf{13.42\!\pm\!0.11}$\\
\bottomrule
\end{tabularx}
\end{table*}

\begin{table*}[!t]
\centering\footnotesize
\setlength{\tabcolsep}{2.5pt}
\renewcommand{\arraystretch}{1.15}
\caption{\textbf{DynaGoal-Deadline results.} SR is in percent and average planning latency in ms. Task-specific execution deadlines and evaluation protocols are specified in Appendix~\ref{sec:seven_extensions}.}
\label{tab:deadline_results}
\begin{tabularx}{\textwidth}{@{}l*{6}{>{\centering\arraybackslash}X}@{}}
\toprule
Method & PushT-deadline & Reacher-deadline & Cube-deadline & TwoRoom-deadline & Avg. SR (\%) & Latency (ms)\\
\midrule
\rowcolor{gray!10}
LeWM~\cite{maes2026lewm} & $5.67\!\pm\!0.58$ & $6.67\!\pm\!1.15$ & $3.33\!\pm\!0.58$ & $33.33\!\pm\!3.21$ & $12.25\!\pm\!0.43$ & $767.50\!\pm\!4.51$\\
\rowcolor{gray!1}
Fast-LeWM~\cite{gao2026fast} & $14.67\!\pm\!2.08$ & $15.33\!\pm\!1.53$ & $6.00\!\pm\!1.00$ & $47.33\!\pm\!4.73$ & $20.83\!\pm\!0.88$ & $404.14\!\pm\!7.44$\\
\rowcolor{gray!10}
Fast-LeWM + SC~\cite{gao2026fast} & $7.00\!\pm\!1.73$ & $4.33\!\pm\!0.58$ & $3.67\!\pm\!1.15$ & $39.67\!\pm\!7.77$ & $13.67\!\pm\!2.13$ & $624.14\!\pm\!16.77$\\
\rowcolor{gray!1}
GC-IDM$^{*}$~\cite{nguyen2026latent} & $8.00\!\pm\!2.65$ & $4.33\!\pm\!1.53$ & $7.00\!\pm\!2.00$ & $30.33\!\pm\!1.53$ & $12.42\!\pm\!0.14$ & $167.12\!\pm\!0.95$\\
\rowcolor{gray!10}
INTACT~\cite{sun2026intact} & $57.00\!\pm\!2.65$ & $82.00\!\pm\!2.65$ & $74.67\!\pm\!2.08$ & $87.67\!\pm\!0.58$ & $75.33\!\pm\!1.18$ & $24.87\!\pm\!0.25$\\
\rowcolor{gray!1}
INTACT Guarded A~\cite{sun2026intact} & $54.67\!\pm\!1.15$ & $60.33\!\pm\!2.31$ & $26.67\!\pm\!2.52$ & $93.33\!\pm\!1.53$ & $58.75\!\pm\!0.90$ & $111.40\!\pm\!1.86$\\
\rowcolor{blue!10}
\textbf{Fast-JEPA (Ours)} & $72.33\!\pm\!3.21$ & $87.00\!\pm\!2.65$ & $93.33\!\pm\!1.15$ & $98.00\!\pm\!1.00$ & $\mathbf{87.67\!\pm\!0.38}$ & $\mathbf{13.75\!\pm\!0.03}$\\
\bottomrule
\end{tabularx}
\end{table*}

\begin{figure*}[!t]
\centering
\includegraphics[width=\textwidth]{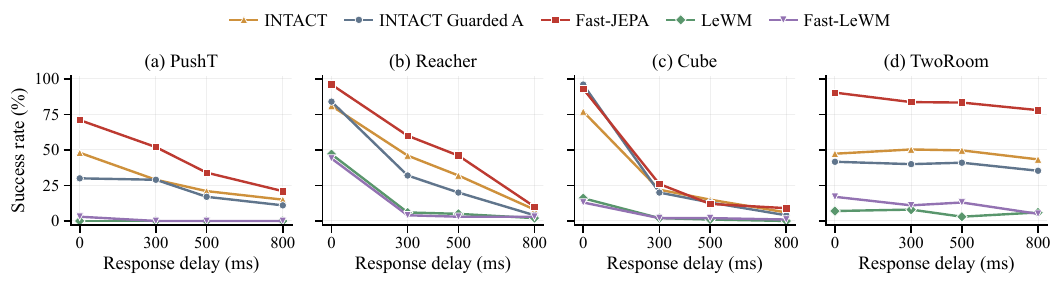}
\caption{\textbf{DynaGoal-Delay results.} Success rates across the four Shift tasks under imposed total response delays of 0, 300, 500, and 800 ms. Each method is evaluated on 100 episodes per task at each delay level.}
\label{fig:dynamic_delay_main}
\end{figure*}

\section{Experiments}
\label{sec:experiments}
We evaluate whether predictive search can deliver
competitive control at low planning latency.
Complete-controller comparisons establish the
success--latency operating point, DynaGoal tests
performance under changing and time-constrained
control conditions, and targeted component analyses
examine the search starting point and candidate-evaluation
cost. Search-budget sweeps and qualitative behavioral
examples assess how these choices affect the available
operating points and selected actions.

\noindent\textbf{Evaluation setting.}
\label{sec:exp_protocol}
We evaluate PushT, Reacher, Cube, and TwoRoom using the
standard native start--goal and success conventions of the
latent-control benchmarks~\cite{maes2026lewm,gao2026fast,sun2026intact}.
The main comparisons average 100 episodes per task over
three independently trained models. We report success rate
and warmed batch-one complete planning latency on an
RTX PRO 6000 GPU. DynaGoal reuses the original-task
models without fine-tuning. Full task, implementation,
and detailed timing protocols appear in
Appendix~\ref{sec:main_evaluation}.

\subsection{Main Results on Four Tasks}
\label{sec:recorded_results}
Fast-JEPA performs predictive search at lower complete-planning latency than INTACT while achieving the highest average success among the locally evaluated methods (Table~\ref{tab:main_results}). It reaches $96.50\%$ at $10.88$\,ms, compared with $95.58\%$ at $26.19$\,ms for INTACT. This corresponds to a $2.41\times$ speedup with a higher average success rate. Relative to INTACT Guarded A, Fast-JEPA maintains a similar success level while reducing planning time from $127.09$ to $10.88$\,ms. This operating point demonstrates that retaining candidate search is compatible with low-latency control.

The task breakdown shows that the aggregate result combines reliable control across distinct environments. Fast-JEPA achieves $100.00\%$ on Cube and $99.67\%$ on TwoRoom, while maintaining high Reacher success. PushT remains more sensitive to the controller choice: Fast-LeWM + SC reaches $92.00\%$, compared with Fast-JEPA's $90.00\%$. Across the four tasks, however, its $84.58\%$ average success and $586.15$\,ms latency contrast with Fast-JEPA's stronger aggregate operating point. 
These results highlight Fast-JEPA's ability to
maintain strong performance while
substantially reducing planning latency across
diverse tasks.

\subsection{Results on DynaGoal}
\label{sec:dynamic}
We evaluate the three DynaGoal protocols introduced in Section~\ref{sec:dynagoal_design}, examining adaptation, timely completion, and robustness to delayed response.

\noindent\textbf{Adaptation to changing conditions.}
Fast-JEPA leads the reported baselines on all four DynaGoal-Shift conditions (Table~\ref{tab:extension_results}), reaching $73.83\%$ average success versus $56.25\%$ for INTACT. The largest separation occurs in TwoRoom, where sustained success under moving geometry reaches $85.33\%$ versus $45.33\%$. Cube also improves from $56.00\%$ to $72.00\%$ under target switching. 
These results extend the original-task performance
to both changing goals and changing environment
geometry, with average planning latency of
$13.42$\,ms across these tasks.

\begin{table*}[!t]
	\centering\footnotesize
	\setlength{\tabcolsep}{3pt}
	\caption{\textbf{Fast-JEPA component ablations.} 
		Fast-JEPA proposal and refinement ablations across
		four tasks. Taskwise and average SR are in percent;
		planning latency is in ms.
		All configurations use matched evaluation settings.
		Case sets and controls appear in
		Appendix~\ref{sec:factorial_protocol}.}
	\label{tab:factorial_plan}
	\begin{tabularx}{\textwidth}{@{}l*{6}{>{\centering\arraybackslash}X}@{}}
		\toprule
		Configuration & PushT & Reacher & Cube & TwoRoom & Avg. SR (\%) & Latency (ms) \\
		\midrule
		\rowcolor{gray!10}
		w/o refinement & $84.00\!\pm\!2.00$ & $92.67\!\pm\!1.53$ & $100.00\!\pm\!0.00$ & $98.67\!\pm\!0.58$ & $93.83\!\pm\!0.63$ & $5.53\!\pm\!0.00$\\
		\rowcolor{gray!1}
		w/o proposal & $45.00\!\pm\!3.46$ & $77.67\!\pm\!5.13$ & $56.00\!\pm\!2.65$ & $94.00\!\pm\!1.00$ & $68.17\!\pm\!0.14$ & $5.27\!\pm\!0.00$\\
		\rowcolor{blue!10}
		\textbf{Fast-JEPA} & $89.00\!\pm\!1.73$ & $96.33\!\pm\!0.58$ & $100.00\!\pm\!0.00$ & $100.00\!\pm\!0.00$ & $\mathbf{96.33\!\pm\!0.52}$ & $10.78\!\pm\!0.06$\\
		\bottomrule
	\end{tabularx}
\end{table*}

\begin{figure*}[!t]
	\centering
	\begin{minipage}{0.495\textwidth}\centering
		\includegraphics[width=\linewidth]{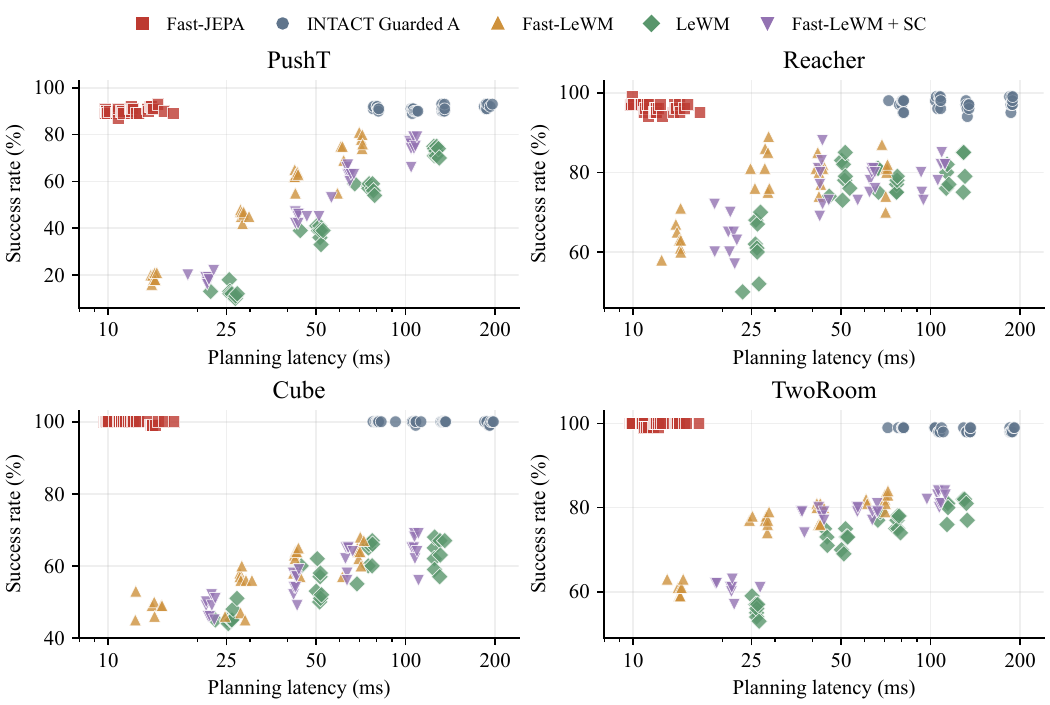}
		\par\small (a) Per-task operating points
	\end{minipage}
	\hfill
	\begin{minipage}{0.495\textwidth}\centering
		\includegraphics[width=\linewidth]{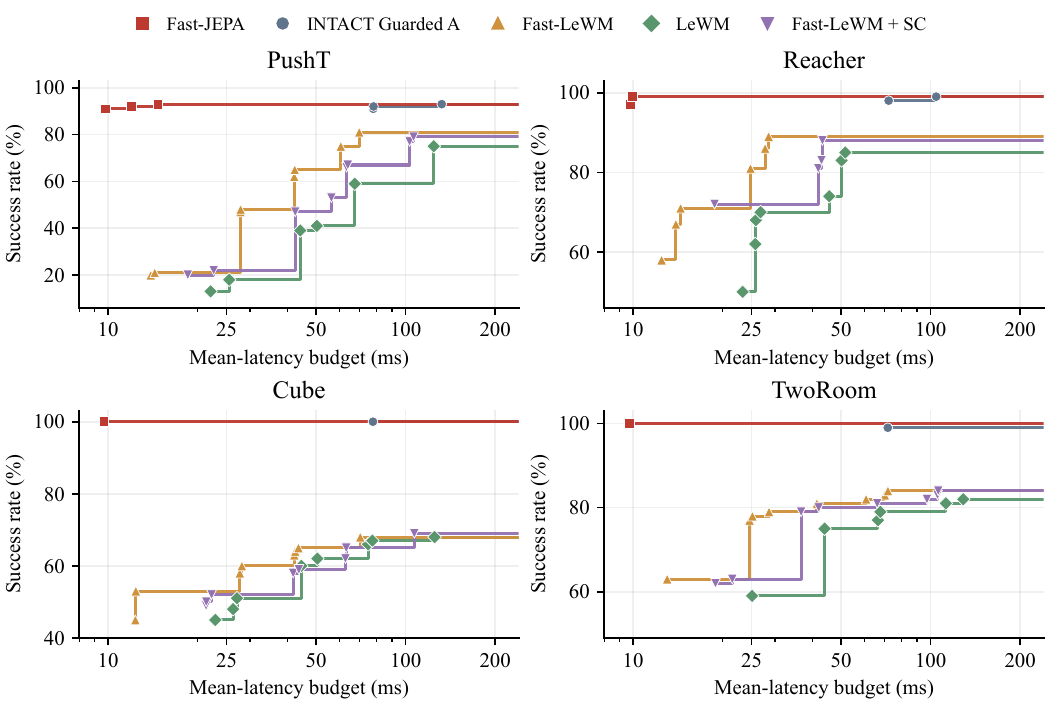}
		\par\small (b) Best observed success within budget
	\end{minipage}
	
	\caption{\textbf{Success--latency comparison across search budgets.}
		(a) Five methods, each evaluated with 32 search configurations,
		show individual success--latency operating points across the
		evaluated configurations. Colors and shapes identify methods.
		(b) Highest observed success within each mean-latency budget,
		computed from the same points as (a).
		Protocols appear in Appendix~\ref{sec:frontier_protocol}.}
	
	\label{fig:frontier_plan}
\end{figure*}

\noindent\textbf{Completion within deadlines.}
Fast-JEPA also leads on every DynaGoal-Deadline task (Table~\ref{tab:deadline_results}), with $87.67\%$ average success compared with $75.33\%$ for INTACT. Cube shows a substantial separation, reaching $93.33\%$ versus $74.67\%$, while TwoRoom reaches $98.00\%$. The combination of high success and $13.75$\,ms average planning latency provides a favorable operating point when planning and action execution share a task deadline.

\noindent\textbf{Robustness to delayed response.}
\label{sec:dynamic_delay_main}
Response delay changes control quality differently across tasks (Figure~\ref{fig:dynamic_delay_main}). Fast-JEPA retains a clear advantage in TwoRoom throughout the delay sweep, while Reacher and Cube success falls sharply as delay grows. PushT shows a more gradual decline. The contrast between tasks indicates that delayed feedback interacts with the control condition, rather than imposing the same performance penalty everywhere. Fast-JEPA achieves higher success than the evaluated alternatives at every tested delay on PushT, Reacher, and TwoRoom. Its sustained lead on TwoRoom highlights the value of proposal-guided predictive refinement when the environment continues to change between observation and action.

The moving-doorway illustration in Figure~\ref{fig:tworoom_delay_schematic} shows how geometry can change between observation and action. Together with Shift and Deadline, these results connect the success--latency comparison to adaptation, timely completion, and robustness under controlled response intervals.

\begin{figure*}[!t]
\centering
\includegraphics[width=\linewidth]{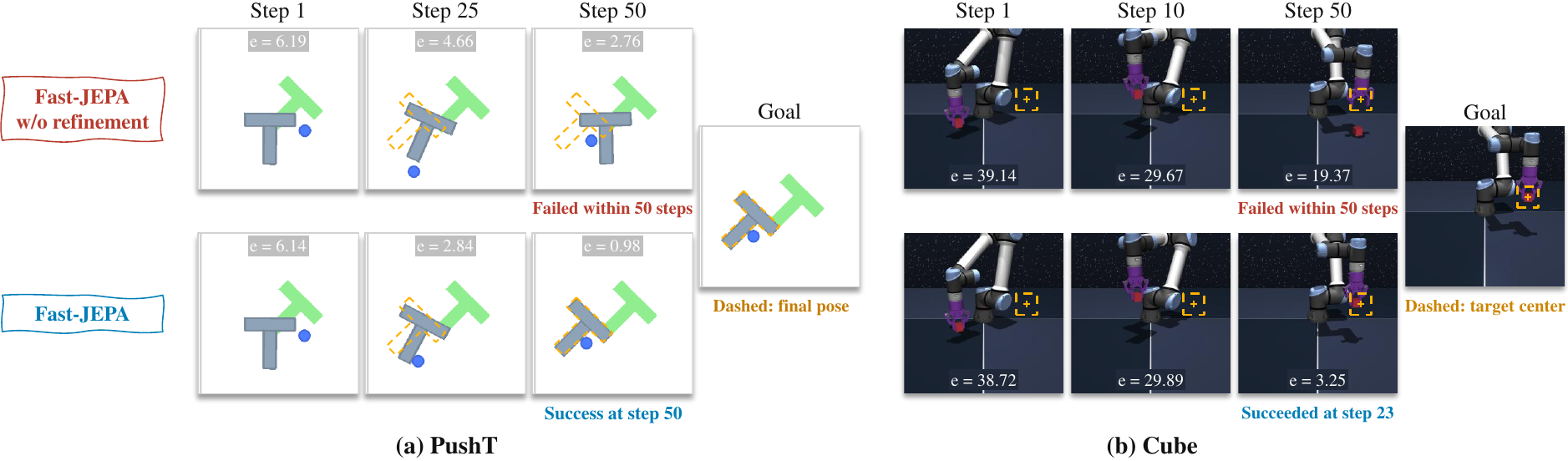}
\caption{\textbf{Predictive refinement rescues two rollouts.} Matched PushT and Cube cases share initial states and goals. Callouts highlight observed alignment and grasp differences. Success and failure are determined within the 50-step evaluation budget. Asterisks mark retained successful terminal observations. Rightmost panels show the evaluated goals.}
\label{fig:qual_plan}
\end{figure*}

\subsection{Ablation Studies}
\label{sec:factorial_plan}

\noindent\textbf{Proposal focuses search.}
Removing the proposal lowers average success from
$96.33\%$ to $68.17\%$ at the same search budget
(Table~\ref{tab:factorial_plan}).
Candidate evaluation alone is therefore insufficient
to recover the complete controller's performance
from broad initialization.
The proposal supplies a productive neighborhood and
a retained plan, allowing refinement to focus on
plausible alternatives.
Effective search thus depends not only on how many
candidates are evaluated, but also on where that
evaluation is concentrated.

\noindent\textbf{Refinement beyond the proposal.}
Without refinement, the controller achieves $93.83\%$
success at $5.53$\,ms, while full Fast-JEPA reaches
$96.33\%$ at $10.78$\,ms.
The gains are task-dependent: refinement improves
PushT and Reacher, while Cube already achieves
$100\%$ success without refinement.
This suggests that additional search is most valuable
when the initial proposal leaves room for improvement.
Proposal-only execution remains a useful low-cost
mode, while refinement provides an opportunity to
improve control through candidate comparison.

\subsection{Success and Latency across Search Budgets}
\label{sec:frontier_plan}

Figure~\ref{fig:frontier_plan}(a) shows a consistent low-latency region for Fast-JEPA across 32 search configurations per task. Its measured planning times span approximately $9.7$--$16.8$\,ms, while INTACT Guarded A spans approximately $71.8$--$197.2$\,ms. Their success ranges overlap on PushT, Reacher, and Cube. On Cube and TwoRoom, every measured Fast-JEPA configuration reaches at least $99\%$ success. The separation in latency therefore persists across the search sweep rather than depending on a single candidate-count or iteration setting.

The empirical budget envelopes in Figure~\ref{fig:frontier_plan}(b) make this separation operational: Fast-JEPA reaches high-success choices at budgets below those required by the measured INTACT Guarded A configurations. Cube and TwoRoom leave little room for additional success gains once Fast-JEPA reaches this region, whereas PushT and Reacher retain more variation across configurations. These task-dependent patterns explain why more search need not produce a proportional improvement. The two panels use the same measured configurations, with envelope construction described in Appendix~\ref{sec:frontier_protocol}. Across candidate counts and refinement depths, Fast-JEPA achieves competitive success within a
substantially lower latency range.

\begin{table}[!t]
	\centering\footnotesize
	\setlength{\tabcolsep}{2.5pt}
	\caption{\textbf{Alternative planning designs.}
		Comparisons of proposal construction and candidate scoring
		under matched search budgets. SR is in percent. Scoring
		and complete planning times are in ms.
		Controls and aggregation appear in
		Appendix~\ref{sec:factorial_protocol}.}
	\label{tab:external_components}
	\begin{tabularx}{\linewidth}{@{}l*{3}{>{\centering\arraybackslash}X}@{}}
		\toprule
		Configuration & Avg. SR (\%) & Scoring (ms) & Latency (ms) \\
		\midrule
		\rowcolor{gray!1}
		GC-IDM proposal & $94.42\!\pm\!0.90$ & $2.40\!\pm\!0.10$ & $132.91\!\pm\!0.21$ \\
		\rowcolor{gray!10}
		Oracle-state proposal & $95.77\!\pm\!0.42$ & $2.35\!\pm\!0.09$ & $781.34\!\pm\!0.08$ \\
		\rowcolor{gray!1}
		SC-augmented scoring & $92.13\!\pm\!0.33$ & $5.85\!\pm\!0.06$ & $14.77\!\pm\!0.09$ \\
		\rowcolor{gray!10}
		Recursive scoring & $96.08\!\pm\!0.76$ & $11.16\!\pm\!0.03$ & $20.54\!\pm\!0.03$ \\
		GC-IDM + Recursive & $90.03\!\pm\!0.61$ & $11.42\!\pm\!0.11$ & $158.30\!\pm\!0.13$ \\
		\rowcolor{blue!10}
		\textbf{Fast-JEPA} & $\mathbf{96.33\!\pm\!0.52}$ & $\mathbf{2.32\!\pm\!0.01}$ & $\mathbf{10.78\!\pm\!0.06}$ \\
		\bottomrule
	\end{tabularx}
\end{table}

\begin{table}[!t]
	\centering\footnotesize
	\setlength{\tabcolsep}{2.5pt}
	\caption{\textbf{Training cost.} Total GPU-hours across four tasks, reported as mean $\pm$ SD over three training seeds.}
	\label{tab:training_cost}
	\begin{tabularx}{\linewidth}{@{}l*{4}{>{\centering\arraybackslash}X}@{}}
		\toprule
		Metric & Fast-JEPA & Fast-LeWM & LeWM & INTACT \\
		\midrule
		\rowcolor{gray!1}
		GPU-hours & $\mathbf{5.95\!\pm\!0.30}$ & $188.05\!\pm\!3.08$ & $18.16\!\pm\!1.06$ & $6.67\!\pm\!0.37$\\
		\bottomrule
	\end{tabularx}
\end{table}

\subsection{Alternative planning designs.}
Table~\ref{tab:external_components} compares alternative
proposal and scoring designs under matched search budgets.
Replacing prediction-guided construction with a GC-IDM
proposal reduces average success from 96.33\% to 94.42\%
and increases planning latency to 132.91 ms.
An oracle-state proposal achieves comparable success
(95.77\%) but requires 781.34 ms due to simulator rollouts,
supporting the efficiency of learned intermediate predictions.
For candidate evaluation, direct terminal scoring achieves
96.33\% success at 10.78 ms, compared with 96.08\% at
20.54 ms for recursive scoring, reducing scoring time
from 11.16 to 2.32 ms.

These comparisons reveal an asymmetry in the role of prediction during planning. Intermediate predictions help construct useful action sequences, whereas recursively predicting intermediate states during candidate evaluation provides little additional benefit under the terminal-goal objective. The consistent advantage of direct scoring across proposals supports allocating predictive computation according to the information required at each planning stage. 

Table~\ref{tab:training_cost} reports training costs
across tasks. Fast-JEPA requires 5.95 GPU-hours,
compared with 6.67 for INTACT, 18.16 for LeWM,
and 188.05 for Fast-LeWM.
Its low-latency planning does not incur
additional training costs.

\subsection{Qualitative Refinement Behavior}
\label{sec:ranking_plan}

The examples in Figure~\ref{fig:qual_plan} show refinement changing the final control outcome. In PushT, the refined trajectory reaches the success criterion while the proposal-only trajectory remains unsuccessful within the evaluation budget. In Cube, the proposal-only trajectory drops the object onto the table, whereas Fast-JEPA retains the grasp and reaches the goal. These cases illustrate how candidate comparison can produce useful behavioral corrections after an informed initial proposal. The paired runs share initial states and goals, and Appendix~\ref{sec:factorial_protocol} records their evaluation interpretation.

\section{Conclusion}

We presented Fast-JEPA, a world-model framework
that advances the success--latency Pareto frontier
through efficient predictive computation.
Intermediate predictions guide action construction,
while direct terminal queries evaluate plans
without recursive rollout.
Fast-JEPA achieves 96.50\% average success at
10.88\,ms across four tasks.
Search-budget sweeps confirm its efficiency across
configurations, while DynaGoal evaluates
out-of-distribution conditions, execution deadlines,
and response delays.

{\small
\bibliographystyle{ieeenat_fullname}
\bibliography{main}
}
\clearpage
\appendix

\section{Implementation and Training}
\label{sec:implementation}
\noindent\textbf{Sequential action proposal.}
The action proposer uses the mean of its conditional action distribution. At chunk $k$,
it consumes $\hat z_0=z_t$ for $k=0$, or
$\hat z_k=W_\psi(z_t,\bar A_{:k})$ for $k>0$, together with the preceding action.
The conditional mean action generator is $G_\phi$ as defined in
Eq.~\eqref{eq:proposal_recursion}.
The prediction after the final chunk is unused and omitted.

\noindent\textbf{Training interface.}
For the three-seed checkpoints, each model family learns its own image
encoder and action interface. Within a checkpoint, current and future images
share a ViT-Tiny encoder~\cite{dosovitskiy2021vit} and an MLP projector with BatchNorm, producing
192-dimensional representations. The outcome prediction is mapped through a
separate prediction projector into this same space. Latent vectors are not
$\ell_2$-normalized before the MSE objective or terminal-distance scoring.
Consequently, equal latent dimension does not make separately trained
checkpoints interchangeable.

\noindent\textbf{Joint training.}
Combining world-model prediction, SiGReg representation regularization, and action
supervision gives
\begin{equation}
 \mathcal L=\mathcal L_{\mathrm{wm}}
 +\lambda_{\mathrm{reg}}\mathcal L_{\mathrm{SiGReg}}
 +\mathcal L_{\mathrm{act}}.
 \label{eq:training_objective}
\end{equation}

The world-model loss in Eq.~\eqref{eq:wm_loss} averages squared error over the first five future chunks,
all latent dimensions, and the batch, with every prediction anchored to the
first observed representation. Future target embeddings retain gradients
in this loss. The objective in Eq.~\eqref{eq:training_objective} uses
SiGReg~\cite{balestriero2025lejepa}, with loss weights
\begin{equation}
 (\lambda_{\mathrm{reg}},\lambda_{\mathrm{next}},
 \lambda_{\mathrm{end}})=(0.02,0.10,0.05).
\end{equation}
The action objective in Eq.~\eqref{eq:action_loss} uses the next-state and
endpoint conditions defined in Eq.~\eqref{eq:action_conditions}. Both conditions
share the same conditional Gaussian predictor.

The action objective in Eq.~\eqref{eq:action_loss} is implemented as a
coordinate-averaged Gaussian negative log-likelihood, with the additive
normalization constant omitted. Both conditions supervise the same
demonstrated action chunk through a shared mean and scale predictor.
The terminal target embedding is detached for the end-conditioned action
term, unlike targets in the world-model prediction loss. Actions are
standardized with the loaded training dataset's column mean and standard
deviation before chunking. No self-consistency loss is added.
The four task configurations use one training epoch, batch size 256, AdamW~\cite{loshchilov2019adamw}
with learning rate $5\times10^{-4}$ and weight decay $10^{-3}$, and
BF16 mixed precision.

\noindent\textbf{CEM implementation.}
The evaluated CEM uses the elite mean and population standard deviation
without smoothing ($\alpha=1$), with standard deviation restricted to
$[10^{-4},10]$ and initial value $\sigma_0=0.25$. Each round overwrites two
candidate slots with the original proposal and the best plan found so far.
The frozen search core scores raw Gaussian actions without clipping them.
Environment action handling is inherited from the benchmark interface.

The update shifts probability toward sequences with lower predicted goal cost.
The evaluated configuration uses $N=64$, $I=2$, and $K=8$. Because two slots
retain existing plans, $NI$ counts scored candidate slots rather than distinct
random alternatives. After the last round, the final distribution mean is
scored once and compared with the best evaluated plan. The lower-cost plan is
returned, so the solve scores $NI+1$ sequences across $I+1$ scoring batches.
This preserves the proposal in predicted cost, without guaranteeing an
improvement in realized control. The selected plan is executed for $L$ actions.

\noindent\textbf{Execution interval.}
The evaluated controller uses $H=5$, $B=5$, and $L=25$: it executes the full
five-chunk plan before refreshing the observation. Static evaluation permits
50 primitive actions and uses a goal 25 dataset steps after the initial image.

\section{Evaluation Protocol and DynaGoal}
\label{sec:main_evaluation}
\subsection{Original-task evaluation}
\noindent\textbf{Task sources and controller implementations.}
PushT follows the Diffusion Policy evaluation suite~\cite{chi2023diffusionpolicy}. Reacher and Cube are drawn from DeepMind Control~\cite{tassa2018dmc} and OGBench~\cite{park2025ogbench}, respectively, using the LeWM setup~\cite{maes2026lewm}. Reacher and Cube use MuJoCo-based simulation~\cite{todorov2012mujoco}. Locally evaluated controllers are LeWM, Fast-LeWM, Fast-LeWM + SC, GC-IDM Direct, INTACT, INTACT Guarded A, and Fast-JEPA. Independent timing uses fixed synthetic inputs.

The original tasks use native success criteria, goal offset 25, and a 50-step execution cap. INTACT and Fast-JEPA use five chunks of five primitive actions and 25-step feedback, with the expert-continuation action history supplied by the benchmark. Training seeds are 0, 42, and 3072, with 100 paired episodes per task and checkpoint. INTACT and INTACT Guarded A share weights. Four-task aggregates first average tasks equally within a trained model, then average the trained-model results. 
All local GPU timing uses an NVIDIA RTX PRO 6000
Blackwell Server Edition.

\begin{table*}[!t]
	\centering\footnotesize
	\setlength{\tabcolsep}{3pt}\renewcommand{\arraystretch}{1.12}
	\caption{\textbf{DynaGoal planning latency (ms).}}
	\label{tab:extension_full_results}
	\begin{tabularx}{\textwidth}{@{}l*{7}{>{\centering\arraybackslash}X}@{}}
		\toprule
		Condition & LeWM & Fast-LeWM & + SC & GC-IDM$^{*}$ & INTACT & Guarded A & Fast-JEPA\\
		\midrule
		\rowcolor{gray!1}
		Shift PushT & $774.70\!\pm\!1.36$ & $408.27\!\pm\!34.49$ & $619.61\!\pm\!48.49$ & $147.68\!\pm\!1.89$ & $24.39\!\pm\!1.62$ & $123.60\!\pm\!6.29$ & $13.58\!\pm\!0.07$\\
		\rowcolor{gray!10}
		Shift Reacher & $754.96\!\pm\!0.99$ & $416.60\!\pm\!0.50$ & $635.59\!\pm\!1.45$ & $174.83\!\pm\!6.62$ & $23.88\!\pm\!0.93$ & $107.56\!\pm\!3.56$ & $13.33\!\pm\!0.21$\\
		\rowcolor{gray!1}
		Shift Cube & $754.76\!\pm\!0.69$ & $402.05\!\pm\!24.43$ & $621.46\!\pm\!49.27$ & $192.75\!\pm\!6.39$ & $25.07\!\pm\!1.86$ & $113.97\!\pm\!7.52$ & $13.32\!\pm\!0.31$\\
		\rowcolor{gray!10}
		Shift TwoRoom & $785.61\!\pm\!8.26$ & $390.79\!\pm\!28.67$ & $597.82\!\pm\!48.54$ & $166.42\!\pm\!3.00$ & $25.14\!\pm\!0.34$ & $116.57\!\pm\!1.72$ & $13.45\!\pm\!0.14$\\
		\rowcolor{gray!1}
		Deadline PushT & $772.84\!\pm\!1.88$ & $410.42\!\pm\!31.24$ & $623.45\!\pm\!50.83$ & $158.08\!\pm\!2.55$ & $26.84\!\pm\!0.18$ & $112.33\!\pm\!8.18$ & $13.96\!\pm\!0.00$\\
		\rowcolor{gray!10}
		Deadline Reacher & $755.95\!\pm\!0.72$ & $419.05\!\pm\!1.52$ & $637.34\!\pm\!2.53$ & $166.42\!\pm\!2.56$ & $23.66\!\pm\!0.02$ & $105.43\!\pm\!0.54$ & $13.30\!\pm\!0.05$\\
		\rowcolor{gray!1}
		Deadline Cube & $755.55\!\pm\!0.66$ & $394.35\!\pm\!28.79$ & $636.20\!\pm\!65.41$ & $192.53\!\pm\!0.76$ & $25.39\!\pm\!1.36$ & $117.30\!\pm\!5.59$ & $14.02\!\pm\!0.10$\\
		\rowcolor{gray!10}
		Deadline TwoRoom & $785.66\!\pm\!16.04$ & $392.74\!\pm\!28.13$ & $599.57\!\pm\!51.08$ & $151.45\!\pm\!0.56$ & $23.58\!\pm\!0.22$ & $110.55\!\pm\!7.82$ & $13.72\!\pm\!0.05$\\
		\bottomrule
\end{tabularx}\end{table*}

\subsection{DynaGoal Benchmark}
\label{sec:seven_extensions}
DynaGoal evaluates original-task models without benchmark-condition training or fine-tuning. OOD refers to changed control conditions and execution constraints, not visual-domain or episode disjointness. Each condition fixes 100 start--goal cases and reuses the three original-task checkpoints. Reproduction requires the same case manifests, checkpoint identities, native success criteria, and timing boundaries.

The four Shift conditions run for 100 primitive steps with 25-step feedback. PushT drifts at 2 pixels/s, reverses direction after step 50, and requires success throughout the final 25 steps. Reacher switches its target after step 75 and requires success strictly before and after the switch. Cube switches the target after step 25 and requires reaching both stages. TwoRoom has an 8-pixel sliding-door amplitude, a 20-second period, and zero phase. Its physical goal remains fixed while the goal image reflects the current door geometry. Success requires the original 16-pixel goal tolerance throughout the final 25 steps.

Deadline conditions use 2.50\,s for PushT, 1.20\,s for Reacher, 1.25\,s for Cube, and 2.50\,s for TwoRoom. All methods use a 50-step execution cap across the four DynaGoal-Deadline tasks. Native step durations are 0.10, 0.04, 0.05, and 0.10\,s for PushT, Reacher, Cube, and TwoRoom. Deadline success requires the first goal hit before cumulative actual planning time plus execution duration exceeds the deadline. In Shift, measured planning delay determines when a plan becomes available in native simulation ticks, with buffered or held actions during the wait. 
Taskwise planning latencies for DynaGoal-Shift and
DynaGoal-Deadline are reported in
Table~\ref{tab:extension_full_results}.
GC-IDM latencies marked $^{*}$ are single-decision
times multiplied by 25 for comparison; this scaling
is not used in deadline evaluation.

\section{Component and Behavioral Analyses}
\label{sec:factorial_protocol}
\label{sec:completed_factorial}
The proposal-only configuration retains the action proposer and its intermediate outcome queries, then executes the proposal without candidate evaluation. Each row aggregates the four tasks within each of three trained models, then reports the mean and sample standard deviation across models, using 100 episodes per task and checkpoint. Qualitative comparisons pair initial states and goals, not identical intermediate-state branches. In the PushT example, native success is first reached at step 23, with the later frame shown for visualization. Both unsuccessful trajectories remain unsuccessful within the 50-step budget.

Tables~\ref{tab:factorial_plan} and~\ref{tab:external_components} report component ablations and alternative planning designs, respectively, with
the complete Fast-JEPA controller as a reference. Every search configuration uses $N=64$, $I=2$, and $K=8$, fixed horizon, feedback, output rules, encoding caches, and unused-terminal pruning. Broad search starts from $\mathcal N(0,I)$ in standardized action space without a retained composed proposal. Proposal-centered search starts from the shared Fast-JEPA proposal with standard deviation 0.25 and retains the proposal and best-so-far slots. This tests the proposal mechanism as a whole, including initialization, retention, and computation cost.

\noindent\textbf{Oracle-state proposal.}
We replace intermediate world-model predictions during
proposal construction with encoded observations obtained
by executing action prefixes in a separate simulator
initialized from the same state.
The action generator, terminal scorer, and CEM settings
remain unchanged.

\section{Search Budgets and Computational Cost}
\label{sec:completed_frontier}
\subsection{Search Budgets and Timing}
\label{sec:frontier_protocol}
\noindent\textbf{Five-method configuration sweep.}
Figure~\ref{fig:frontier_plan}(a) uses 32 combinations of $N\in\{32,48,64,96,128,192,256,384\}$ and $I\in\{1,2,3,5\}$, with $K=N/8$. Fast-JEPA, INTACT Guarded A, LeWM, Fast-LeWM, and Fast-LeWM + SC each use a checkpoint trained with seed 0 on each original task. Every configuration is evaluated on 100 fixed cases per task under the 50-step execution cap. Planning covers 25 primitive actions with 25-step feedback. Checkpoints, cases, and method-specific execution settings remain fixed across configurations. The plotted values are success rates for the fixed case set and mean planning times.

\noindent\textbf{Empirical budget envelope.}
For method $m$, task $t$, and latency allowance $b$, Figure~\ref{fig:frontier_plan}(b) plots
\begin{equation}
E_{m,t}(b)=\max_{c\in\mathcal C:\,\bar L_{m,t,c}\le b}S_{m,t,c},
\end{equation}
where $\mathcal C$ is the same 32-configuration grid used in panel (a), $S$ is observed success, and $\bar L$ is mean complete planning latency. The envelope is undefined when no measured configuration fits the allowance. Curves change only when an additional configuration becomes feasible and improves observed success. They use no interpolation between configurations. The envelope selects and reports on the same evaluation cases, so it is descriptive rather than an independently validated configuration-selection result. A mean-latency allowance does not guarantee that every call meets a hard deadline.

\subsection{Computational scaling}
\label{sec:scaling_protocol}
\begin{figure}[h]
\centering
\includegraphics[width=\linewidth]{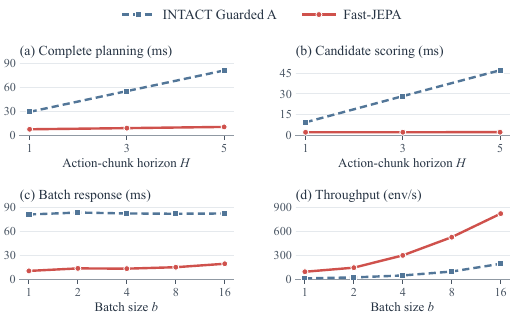}
\caption{\textbf{Planning cost scaling.} Complete response and scoring are measured separately.}
\label{fig:scaling_plan}
\end{figure}

Figure~\ref{fig:scaling_plan} varies horizon $H\in\{1,3,5\}$ at batch one and batch size $b\in\{1,2,4,8,16\}$ at $H=5$, holding search at $64\times2$/top8. Available points include all four tasks and three trained models. Full response uses 1,000 synchronized calls per cell, while scoring uses 40 separate profiles. Throughput is total environments divided by batch response time.
\section{DynaGoal-Delay Protocol}
\label{sec:dynamic_delay_protocol}
DynaGoal-Delay evaluates controller robustness on the four DynaGoal-Shift tasks defined in Appendix~\ref{sec:seven_extensions}, retaining their task dynamics and success criteria. Each controller is evaluated on 100 episodes per task under imposed total response delays of 0, 300, 500, and 800\,ms. These values set the response interval rather than adding to measured computation time. During this interval, the environment continues to advance using buffered actions from the previous plan. When no buffered actions remain, the last action is held. The new plan is executed once available. All methods use the same delay-handling mechanism.

\end{document}